\documentclass{article}

\usepackage{PRIMEarxiv}

\usepackage[utf8]{inputenc} % allow utf-8 input
\usepackage[T1]{fontenc}    % use 8-bit T1 fonts
\usepackage{hyperref}       % hyperlinks
\usepackage{url}            % simple URL typesetting
\usepackage{booktabs}       % professional-quality tables
\usepackage{amsfonts}       % blackboard math symbols
\usepackage{nicefrac}       % compact symbols for 1/2, etc.
\usepackage{microtype}      % microtypography
\usepackage{lipsum}
\usepackage{fancyhdr}       % header
\usepackage{graphicx}       % graphics
\graphicspath{{media/}}     % organize your images and other figures under media/ folder

\usepackage{amssymb}
\usepackage{amsmath}
\usepackage{mathtools}
\usepackage{mathrsfs}
\usepackage{hyperref}
\usepackage{algorithm}
\usepackage{algorithmic}
\usepackage{makecell}
\usepackage{enumitem}
\usepackage{tabularx}
\usepackage[title]{appendix}
\usepackage{amsthm}
\usepackage[capitalize]{cleveref}
\usepackage{natbib}

\title{Fractional-order Jacobian Matrix Differentiation and Its Application in Artificial Neural Networks
\thanks{
This research was funded by the Yunnan University Graduate Research Innovation Fund Project.} 
}

\author{
  Xiaojun Zhou\textsuperscript{1,2}, Chunna Zhao\textsuperscript{1}, Yaqun Huang\textsuperscript{1}, Chengli Zhou\textsuperscript{1}, Junjie Ye\textsuperscript{1}, Kemeng Xiang\textsuperscript{1} \\
  \textsuperscript{1}School of Information Science and Engineering, Yunnan University, Kunming \\
  \textsuperscript{2}Lijiang Normal University, Lijiang\\
  \texttt{zxjssldqm@163.com}
}

\begin{document}
\maketitle

\begin{abstract}
% \lipsum[1]
Fractional-order differentiation has many characteristics different from integer-order differentiation. These characteristics can be applied to the optimization algorithms of artificial neural networks to obtain better results. However, due to insufficient theoretical research, at present, there is no fractional-order matrix differentiation method that is perfectly compatible with automatic differentiation (Autograd) technology. 
Therefore, we propose a fractional-order matrix differentiation calculation method. This method is introduced by the definition of the integer-order Jacobian matrix. We denote it as fractional-order Jacobian matrix differentiation (${{\bf{J}}^\alpha }$). Through ${{\bf{J}}^\alpha }$, we can carry out the matrix-based fractional-order chain rule. Based on the Linear module and the fractional-order differentiation, we design the fractional-order Autograd technology to enable the use of fractional-order differentiation in hidden layers, thereby enhancing the practicality of fractional-order differentiation in deep learning. In the experiment, according to the PyTorch framework, we design fractional-order Linear (FLinear) and replace nn.Linear in the multilayer perceptron with FLinear. Through the qualitative analysis of the training set and validation set $Loss$, the quantitative analysis of the test set indicators, and the analysis of time consumption and GPU memory usage during model training, we verify the superior performance of ${{\bf{J}}^\alpha }$ and prove that it is an excellent fractional-order gradient descent method in the field of deep learning.
\end{abstract}

% keywords can be removed
\keywords{Fractional-order optimization methods, Fractional-order matrix differentiation, Fractional-order Jacobian matrix, Fractional-order gradient descent method, Artificial neural networks}

\section{Introduction}
\label{sec1}
Integer-order matrix differentiation is an important theoretical basis for gradient calculation in artificial neural networks (ANN). 
Based on it, scholars have proposed many high-performance deep learning optimizers, which have promoted the development of ANN technology. Deep learning optimization algorithms have always been improved from the perspectives of first-order and second-order differentiations.  
However, with the development of fractional-order differentiation, some scholars began to study deep learning optimization algorithms from the perspective of fractional-order differentiation. These methods can be collectively referred to as fractional-order gradient descent (FGD) methods.

FGD methods have different characteristics from integer-order methods in theory. These characteristics include having built-in advanced learning rate scheduling strategies and being less likely to get stuck in saddle points (See Appendix \ref{appendix1}), as well as having built-in regularization capabilities (Proof in Appendix \ref{appendix2}). Due to these advantages, scholars have attempted to introduce fractional calculus into deep learning and have achieved promising results \cite{yang2021robust, roy2016fractional, yang2025image}. However, in ANNs, data are represented in tensor form. Consequently, when fractional differentiation is extended from scalars to tensors, the fractional matrix differentiation within the hidden layers of ANNs cannot be computed in the same manner as integer-order differentiation, making the matrix-form chain rule inapplicable. As a result, fractional-order differentiation exhibits poor compatibility with integer-order automatic differentiation (Autograd) frameworks.

Currently, there are three approaches to applying FGD methods in ANNs:
(1) Start from the perspective of symbolic differentiation to study FGD methods. To avoid calculating fractional-order matrix differentiation and the subsequent chain calculations in ANNs, they only use fractional-order matrix differentiation on the Mean Squared Error Loss function (MSELoss). That is, they use the fractional-order power function differentiation formula to solve the differentiation of child nodes in the root node, then perform back-propagation based on the chain rule, and participate in the calculation of gradients of intermediate and leaf nodes. In fact, the differentiations of intermediate and leaf nodes are obtained using integer-order Autograd technology. No matter how the MSELoss is transformed, essentially, these are methods that can only use fractional-order differentiation at the output layer \cite{wang2022study,khan2019fractional,chen2017study,zhou2025fractional,lou2022variable}.
(2) Transform the Grünwald-Letnikov (G-L) formula and use Autograd to make the gradient solution become the nonlinear accumulation of gradients at different time nodes. Essentially, they are integer-order optimization methods with special momentum and special learning rates \cite{zhou2023deep,yu2022fractional,kan2021convolutional}.
(3) Abandon the chain rule and use a specific symbolic differentiation method to solve gradients in specific ANNs based on Leibniz's rule \cite{xie2023fractional}.

The above methods conclude that fractional-order differentiation is superior to integer-order differentiation through theoretical analysis and experimental comparison. However, in practical application scenarios, the defects of these methods are also obvious. On the one hand, scholars have ignored the combination of relevant methods with mainstream deep learning frameworks and have not designed fractional-order deep learning optimizers from the perspective of Autograd technology. This may affect their further promotion in the field of deep learning optimization methods. On the other hand, many fractional-order parameter matrix gradients are not derived from Jacobian matrix differentiation like integer-order matrices. Many of them are computationally complex during the solution process, and when the formulas are transformed, the basis of fractional-order matrix differentiation is weakened to varying degrees \cite{elnady2025comprehensive}. In short, the fractional-order deep learning optimizers proposed based on existing FGD methods are not sufficient to comprehensively surpass the numerous integer-order deep learning optimizers.

In addition, scholars have conducted convergence analysis on the proposed FGD methods, and most of them emphasize faster convergence rate on the training set than the corresponding integer-order methods. However, the existing FGD methods and their variants are improvements based on existing first-order optimization algorithms, and their convergence performance is the same as that of the corresponding integer-order optimization algorithms.

Since there is a linear mapping on the gradient matrix between some existing FGD methods and integer-order optimization methods or their variants, adjusting their fractional-order orders can ultimately be regarded as adjusting the learning rate. This further affects the engineering application of FGD methods. Therefore, in order for FGD methods to have long-term development and application in deep learning, we need to re-examine FGD methods and redefine them according to the research ideas of Autograd technology and integer-order matrix differentiation. In this process, the biggest problem faced by FGD methods is that fractional-order differentiation cannot be applied to the hidden layers of ANNs \cite{zhou2025improved,chen2020adaptive,zhu2018fractional,chen2022fractional,wahab2022performance,viera2022artificial,joshi2023survey}.

\begin{figure*}[h]
    \centering
    \includegraphics[width=\textwidth]{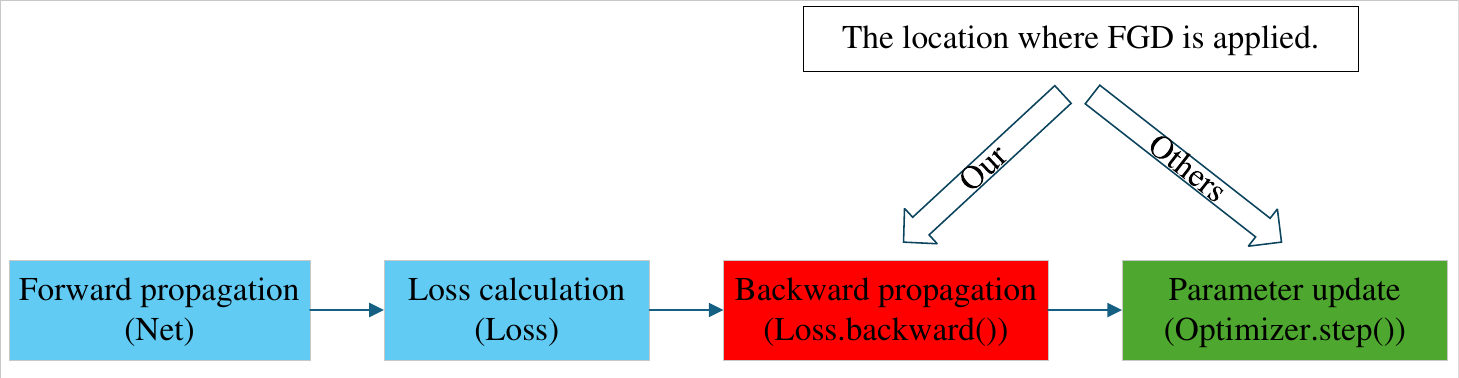}
    \caption{Location differences in operational scope between the proposed method and existing approaches in PyTorch‌.}
    \label{fig1}
\end{figure*}

Thus, This paper proposes fractional-order Jacobian matrix differentiation ${{\bf{J}}^\alpha }$ based on the definition of integer-order Jacobian matrices and fractional-order symbolic differentiation. 
In the experiment, combined with the mainstream deep learning framework PyTorch, the Linear module is reconstructed into a fractional-order Linear module. In fact, this paper preliminarily proposes fractional-order Autograd technology. By similarly reconstructing other modules in torch.nn, we can also establish a complete ecosystem of fractional-order Autograd technology. 
The biggest difference between the work in this paper and previous FGD methods is that through calculating the differentiations of each node on the computation graph, the fractional-order gradient is finally obtained. Unlike existing FGD methods that make modifications in Optimizer.step(). In fact, this paper performs fractional-order gradient calculation in Loss.backward() (see Figure \ref{fig1}).

In specific details, we extract the fractional-order differentiation matrix required for solving the gradient matrix from ${{\bf{J}}^\alpha }$. Its size is the same as that of the integer-order differentiation matrix. In contrast, for integer-order cases, only multiple completely identical differentiation matrices can be defined from the definition of ${{\bf{J}}^\alpha }$. 
When $\alpha = 1.0$, the block matrix obtained is the same as the differentiation matrix obtained in traditional integer-order cases, that is, only one differentiation matrix can be obtained. On the contrary, fractional-order cases can define multiple different differentiation matrices. 
Since scalar-based calculations are impractical in ANNs, we need to convert the calculations of each differentiation matrix in ${{\bf{J}}^\alpha }$ into matrix-based calculations. 
The size of the differentiation matrix defined by ${{\bf{J}}^\alpha }$ is related to the column size of the parameter matrix (usually the weight matrix in this paper), and each parameter matrix has a different size. Therefore, each parameter matrix can select a different number of differentiation matrices for gradient calculation.
In matrix-based algorithm design, to reduce the complexity of the problem, in this paper, we only select the first block matrix in ${{\bf{J}}^\alpha }$ (i.e., the first block matrix on the diagonal) as the fractional-order differentiation matrix to participate in the gradient calculation of the weight matrix in the leaf node. 
Compared with previous research and applications, the fractional-order weight gradient matrix in this paper is not a linear mapping of the integer-order gradient matrix, nor is it equivalent to a variant of the gradient descent (GD) method. On the contrary, this paper's method fully utilizes the learning rate scheduling strategy and regularization characteristics of fractional-order optimization.

In summary, in this paper:
(1) We expound on the theoretical method of ${{\bf{J}}^\alpha }$.
(2) We present a fractional-order linear function differentiation formula suitable for the field of deep learning.
(3) In a given multilayer perceptron (MLP), following the idea of Autograd technology, we combine the fractional-order linear function differentiation formula with ${{\bf{J}}^\alpha }$ and present a gradient solution algorithm based on matrix calculations.
(4) We conduct an experiment on a regression task based on MLP. In the experiment, using the deep learning framework PyTorch and combining Autograd technology, we reconstruct the Linear module, adding a fractional-order backward function and a fractional-order matrix differentiation solver in it. In fact, we design a fractional-order Linear module (FLinear) in PyTorch. And FLinear is a module that contains Linear. 
In the experiment, we replace the Linear layer in the ANN with the FLinear layer, enabling the model to use the matrix-based fractional-order chain rule during back-propagation. Through performance analysis of different orders (when $\alpha = 1.0$, FLinear is the same as Linear), ${{\bf{J}}^\alpha }$ shows superiority.

In deep learning, the method proposed in this paper enables fractional-order matrix differentiation to be applied inside ANNs, not just at the output layer. This allows the FGD method to no longer rely on integer-order Autograd technology and makes it a method that includes integer-order differentiation. Therefore, the contributions of this paper can be summarized as follows:

\begin{itemize}
\item Based on integer-order Jacobian matrix differentiation, ${{\bf{J}}^\alpha }$ is proposed, and its block matrix characteristics are studied. The characteristics of the first block matrix of the fractional-order and the integer-order are compared and analyzed experimentally.
\item The proposed fractional-order matrix differentiation method can avoid explicit operations on matrix elements, thus improving the computational efficiency of the FGD method.
\item This paper focuses on applying the FGD method in the hidden layers of ANNs and explains and experimentally analyzes the FGD method and its variants from a mathematical theory perspective. Closely integrated with Autograd technology, its application ability is generalized. This makes the FGD method no longer a method that only uses fractional-order differentiation at the root node.
\item Since this paper studies the FGD method based on the idea of Autograd technology, in fact, we have found a breakthrough application for fractional-order automatic differentiation (FAutograd) technology. This makes the FGD method and its variants no longer a linear or simple non-linear transformation of the integer-order weight matrix gradient in Optimizer.Step(). Instead, it is an operation on the computation graph in Loss.backward().
\item Through extensive research on the application of the FGD method in ANNs, this paper re-states and defines the application rules of the FGD method and clarifies the non-necessity of proving the convergence of the FGD method.
\end{itemize}

The remaining parts of this paper are organized as follows: Section \ref{sec2} introduces deep learning optimization strategies and the current research status of the FGD method, as well as the application method of fractional-order differentiation in MLPs. Section \ref{sec3} proposes the method in this paper, including ${{\bf{J}}^\alpha }$, the fractional-order differentiation formula in ANNs, and the calculation of the fractional-order gradient matrix. Section \ref{sec4} is about the experiments. Section \ref{sec5} is the summary.

\section{Related Work and Motivation}
\label{sec2}
In this section, we will introduce existing deep learning strategies and FGD methods, and expound on their relationship. On this basis, we will introduce in detail the application of FGD methods in ANNs. Finally, we will explain the motivation for the work in this paper.

\subsection{Deep Learning Optimization Strategies and Fractional-order Gradient Descent}
\label{sec21}
In the parameter optimization methods of neural networks, integer-order deep learning optimization algorithms are divided into first-order and second-order optimization methods. The FGD method discussed in this paper is an extension based on first-order optimization methods. Therefore, this section first reviews first-order optimization methods. Then, it explores the learning rate scheduling strategies in deep learning optimization methods. Finally, it re-examines the current research status of existing FGD methods and summarizes them.

\textbf{Integer-order Deep Learning Optimization Methods:}
In the integer-order deep learning framework PyTorch, the Stochastic Gradient Descent (SGD) optimizer is an integrated solution. It includes the SGD method \cite{ruder2016overview}, the Momentum method \cite{qian1999momentum}, and the Nesterov Accelerated Gradient (NAG) method \cite{1983A}. These three optimization methods are achieved by adjusting the parameters of the optimizer. Among them, the SGD method is the most basic optimization method. Due to its inherent problems \cite{sutton1986two}, scholars proposed the Momentum and NAG methods to improve the SGD optimizer. The Momentum method uses past gradient information, while the NAG method uses future gradient information to optimize the parameters being iterated. In practical application scenarios, the SGD optimizer is usually combined with them to accelerate model convergence and help the model escape local optima. 
Another major category of first-order optimization methods is adaptive learning rate optimizers. These optimizers include Adagrad \cite{duchi2011adaptive}, Adadelta \cite{zeiler2012adadelta}, RMSprop \cite{tieleman2012lecture}, and Adam \cite{kingma2014adam}, etc. These adaptive learning rate optimizers use the information of the first-order or second-order moments of the gradient, or both, and achieve iterative updates of the parameter matrix through a non-linear transformation of the gradient matrix. They are equivalent to improving the step size of the gradient matrix in each iteration.

In fact, adaptive learning rate optimizers are equivalent to giving each element of the gradient matrix a separate learning rate, which is a non-linear transformation of the matrix.

\textbf{Learning Rate Scheduling Strategies:}
In addition to using the first-order and second-order moments of the gradient, another important method in the training of ANNs is the learning rate scheduling strategy. During the iteration process, these strategies dynamically adjust the learning rate to increase the step size of each iteration. Such techniques include learning rate decay, cosine annealing \cite{loshchilov2016sgdr}, and warm-up \cite{gotmare2018closer}. In fact, they are scalar multiplications of the matrix, which are linear transformations of the matrix.

That is to say, there is an important difference between integer-order deep learning optimization methods and learning rate scheduling strategies at the matrix level. In terms of the transformation of the gradient matrix, integer-order deep learning optimization methods are non-linear transformations. On the contrary, learning rate adjustment strategies are linear transformations. However, from the longitudinal process of iteration, most learning rate decay methods, cosine annealing, and warm-up can be regarded as non-linear transformations on the gradient matrix. Based on these discussions, when introducing existing FGD methods, we can find their inherent characteristics and determine whether the gradient matrices are linear transformations or non-linear transformations in Optimizer.step().

\textbf{Existing FGD Methods:}
In the research of ANNs, FGD methods use fractional-order differentiation and integer-order optimization techniques to develop new deep learning optimization techniques. Some of these methods use integer-order gradients for solving. References \cite{zhou2023deep,yu2022fractional,kan2021convolutional} are examples. By transforming the G-L formula and combining it with automatic differentiation technology, they finally derive the fractional-order gradient. Essentially, they can be classified as a learning rate scheduling strategy or a momentum method with a sliding window. In fact, they can also be regarded as a special integer-order momentum method. Moreover, they are based on numerical differentiation and are not convenient for compatibility with Autograd technology.
Similarly, references \cite{shin2023accelerating,wang2017fractional} also use numerical differentiation. Although the Caputo definition is used and fractional-order differentiation is specifically applied to MSELoss at the root node, from the perspective of ANNs, this is equivalent to using fractional-order differentiation at the output layer while maintaining integer-order differentiation in the hidden layers. 
Another category of FGD methods is based on symbolic differentiation for solving. In particular, the fractional-order differentiation formula of quadratic functions is used for solving, that is, fractional-order differentiation is applied to MSELoss \cite{khan2019fractional}. Reference \cite{wang2022study} transforms the quadratic function before applying fractional-order differentiation to alleviate the gradient explosion problem, it caused by the loss of gradient direction due to the absolute value symbol. Reference \cite{xie2023fractional} also uses symbolic differentiation, but it abandons the chain rule and uses Leibniz's rule in specific ANNs. In fact, by transforming the differentiation formula, the use of absolute values is avoided. However, including reference \cite{zhou2025fractional}, these FGD methods using symbolic differentiation may have a linear mapping with integer-order optimization methods. This linear mapping can be achieved by adjusting the learning rate in integer-order deep learning optimizers. Even if the gradient matrix of some of these methods is no longer a linear mapping relationship after formula transformation, essentially, they operate on the basis of the integer-order gradient matrix, rather than directly solving the fractional-order gradient matrix of the leaf nodes on the computation graph.
The advantage of FGD methods based on symbolic differentiation is that they can use the idea of Autograd technology. However, generally speaking, these methods may have two problems: (1) The fractional-order differentiation formula is heavily modified to avoid complex fractional-order matrix differentiation. (2) When performing the chain rule in matrix form, to solve row-column mismatch between two differentiation matrices in matrix multiplication, matrix multiplication is changed to the Hadamard product.

\subsection{Four Cases of Solving Weights with Fractional-order Differentiation in Multilayer Perceptrons}
\label{sec22}
\begin{figure*}[h]
    \centering
    \includegraphics[width=0.6\hsize]{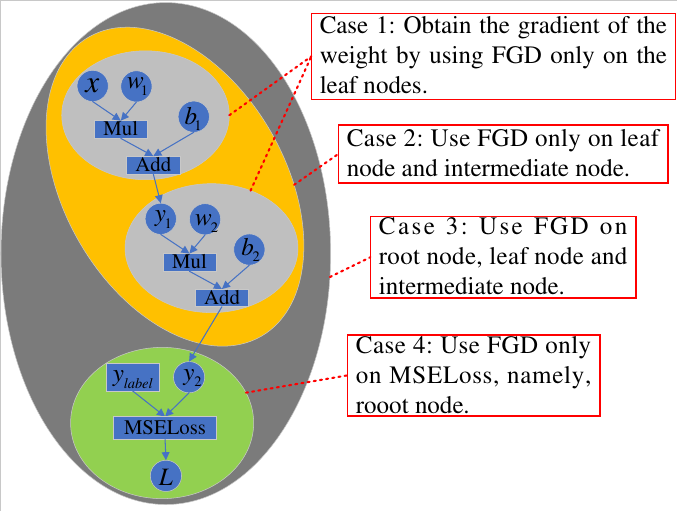}
    \caption{Computational graph of ANN and 4 cases of FGD in back-propagation‌.}
    \label{fig2}
\end{figure*}

To further elaborate on the current research status of the FGD method in ANNs, we use a MLP with two linear functions and a quadratic function as the ANN, and introduce four cases of using the FGD method. This ANN can be expressed as \cref{eq1}, and the corresponding computation graph is shown in Figure \ref{fig2}. Specifically, to simplify the description of the example problem and meet the requirements of hyperparameters for subsequent experiments, we first assume that the variables and parameters in \cref{eq1} are all scalars, and there is no activation function after each linear function.

\begin{equation}
  \label{eq1}
  \left\{
  \begin{aligned}
    y_1 &= x w_1 + b_1 \\
    y_2 &= y_1 w_2 + b_2 \\
    L   &= (y_2 - y_{\text{label}})^2
  \end{aligned}
  \right.
\end{equation}
where $x$ represents the input data, $\{w_1, w_2\}$ are the weights, $\{b_1, b_2\}$ are the biases, $y_{\text{label}}$ is the label, and $L$ is the value of the loss function ($Loss$).

In Figure \ref{fig2}, through the chain rule and fractional-order differentiation, four methods for calculating the gradients of leaf nodes can be summarized and defined. Taking $w_1$ as an example, the definitions of its four gradient calculation methods are as follows:

\noindent Case 1:
\begin{equation}
  \label{eq2}
\nabla _{{w_1}}^{(1,1,\alpha) }L({w_1};x,{b_1},{y_1},{y_2}) \coloneqq  \frac{{\partial L}}{{\partial {y_2}}}\frac{{\partial {y_2}}}{{\partial {y_1}}}\frac{{{\partial ^\alpha }{y_1}}}{{\partial {{({w_1})}^\alpha }}}
\end{equation}
Here, $\nabla _{{w_1}}^{(1,\alpha,\alpha) }L({w_1};x,{b_1},{y_1},{y_2})$ means that both integer-order and fractional-order are used when calculating the gradient of $w_1$. $\frac{{{\partial ^\alpha }{y_1}}}{{\partial {{({w_1})}^\alpha }}}$ represents the $\alpha$-th order partial derivative of $y_1$ with respect to $w_1$.

\noindent Case 2:
\begin{equation}
  \label{eq3}
\nabla _{{w_1}}^{(1,\alpha,\alpha) }L({w_1};x,{b_1},{y_1},{y_2}) \coloneqq \frac{{\partial L}}{{\partial {y_2}}}\frac{{{\partial ^\alpha }{y_2}}}{{\partial {{({y_1})}^\alpha }}}\frac{{{\partial ^\alpha }{y_1}}}{{\partial {{({w_1})}^\alpha }}}
\end{equation}

\noindent Case 3:
\begin{equation}
  \label{eq4}
\nabla _{{w_1}}^\alpha L({w_1};x,{b_1},{y_1},{y_2}) \coloneqq \frac{{{\partial ^\alpha }L}}{{\partial {{({y_2})}^\alpha }}}\frac{{{\partial ^\alpha }{y_2}}}{{\partial {{({y_1})}^\alpha }}}\frac{{{\partial ^\alpha }{y_1}}}{{\partial {{({w_1})}^\alpha }}}
\end{equation}
Here, $\nabla _{{w_1}}^\alpha L({w_1};x,{b_1},{y_1},{y_2})$ represents the fractional-order gradient of $L$ with respect to $w_1$.

\noindent Case 4:
\begin{equation}
  \label{eq5}
\nabla _{{w_1}}^{(\alpha ,1,1)}L({w_1};x,{b_1},{y_1},{y_2}) \coloneqq \frac{{{\partial ^\alpha }L}}{{\partial {{({y_2})}^\alpha }}}\frac{{\partial {y_2}}}{{\partial {y_1}}}\frac{{\partial {y_1}}}{{\partial {w_1}}}
\end{equation}

Among the four gradient calculation formulas, Case 4 corresponds to some characteristics of existing FGD methods \cite{wang2022study,khan2019fractional,chen2017study,zhou2025fractional,lou2022variable}. In back-propagation, Case 4 is a method used only at the output layer. 
Conversely, Cases 1, 2, and 3 all involve the problem of fractional-order matrix differentiation in the internal calculation of neural networks. That is, according to the application of existing FGD methods in Sections \ref{sec1} and \ref{sec21}, in the hidden layers of ANNs, Case 4 is a method that must rely on integer-order differentiation to obtain parameter gradients. It is not independent of integer order, and its so-called fractional order does not truly incorporate integer order.

\subsection{Motivation}\label{sec23}

Why can't existing FGD methods be applied to the hidden layers of ANNs? In many studies on fractional-order differentiation, variables and parameters are assumed to be scalars. However, in actual engineering applications, the data, weights, and biases being trained are tensors \cite{elnady2025comprehensive,CAO2023113881,WU2025102947,YAN2023277}. 
In back-propagation, the multiplication between tensors must follow the rules of matrix multiplication. That is, when two differentiation matrices are multiplied, their row and column dimensions must be equal. 
In integer-order matrix differentiation, as long as the matrices in forward propagation satisfy the operation rules of matrix multiplication, the differentiation matrices obtained through the integer-order matrix differentiation solution rules in back-propagation also satisfy the matrix multiplication rules. 
But when we introduce the integer-order calculation idea (such as the Autograd technology of the deep learning framework PyTorch) into fractional-order and use fractional-order differentiation to solve the gradients of matrix functions, we cannot obtain a differentiation matrix that satisfies the matrix multiplication rules. 
In deep learning optimizers, this differentiation matrix needs to participate in chain-based solutions to obtain the gradient matrix. Therefore, we need to study a kind of fractional-order matrix differentiation method so that the obtained differentiation matrix satisfies the matrix-based chain rule.

\section{Method}\label{sec3}

This section systematically introduces ${{\bf{J}}^\alpha }$. Meanwhile, based on previous studies, a fractional-order differentiation formula suitable for the Autograd technology approach is presented. On this basis, the implicit calculation method of the fractional-order gradient matrix is expounded.

\subsection{Fractional-order Jacobian Matrix Differentiation}\label{sec31}

Based on the discussion of the motivation in Section \ref{sec23}, when data and parameters change from scalars to tensors during back-propagation, the defects of existing FGD methods can be known. To solve this problem, we need to make the sizes of the fractional and integer-order differentiation matrices the same. At the same time, when the fractional order $\alpha = 1.0$, the obtained differentiation matrix should be equal to that obtained by the integer-order method. For this purpose, based on the definition of the integer-order Jacobian matrix, we use fractional-order symbolic differentiation on the matrix function to obtain the fractional-order Jacobian matrix. Finally, a block matrix that satisfies the matrix multiplication rules during back-propagation is extracted from the fractional-order Jacobian matrix. The details are as follows:

\textbf{Definition 1.} \label{def1} Let the matrix function ${\mathbf{F(X)}} \in {\mathbb{R}^{p \times q}}$, and the matrix ${\bf{X}} \in {\mathbb{R}^{m \times n}}$. Their vectorizations are defined as

\begin{equation}
\label{eq6}
vec({\bf{F(X)}}) \coloneqq {\left[ {{f_{11}}({\mathbf{X}}), \cdots, {f_{p1}}({\mathbf{X}}), \cdots, {f_{1q}}({\mathbf{X}}), \cdots, {f_{pq}}({\bf{X}})} \right]^{\mathbf{T}}} \in {\mathbb{R}^{pq}}
\end{equation}

\begin{equation}
\label{eq7}
vec({\mathbf{X}}) \coloneqq {\left[ {{x_{11}}, \cdots, {x_{m1}}, \cdots, {x_{1n}}, \cdots, {x_{mn}}} \right]^{\mathbf{T}}} \in {\mathbb{R}^{mn}}
\end{equation}

Then the fractional-order Jacobian matrix of the matrix function $\mathbf{F(X)}$ is defined as

\begin{equation}
\label{eq8}
\mathbf{{J}^\alpha } = 
\nabla _{{{(vec{\mathbf{X}})}^{\mathbf{T}}}}^\alpha {\mathbf{F(X)}} \coloneqq \frac{{{\partial ^\alpha }vec{\mathbf{F(X)}}}}{{\partial {{({{(vec{\mathbf{X}})}^{\mathbf{T}}})}^\alpha }}} \in {\mathbb{R}^{pq \times mn}}
\end{equation}
Its specific representation is \cref{eq9}.

{\footnotesize
\begin{equation}
\label{eq9}
{{\bf{J}}^\alpha } = \left[ \begin{array}{c}
\frac{{{\partial ^\alpha }{f_{11}}}}{{\partial {{({{(vec{\bf{X}})}^{\bf{T}}})}^\alpha }}}\\
 \vdots \\
\frac{{{\partial ^\alpha }{f_{p1}}}}{{\partial {{({{(vec{\bf{X}})}^{\bf{T}}})}^\alpha }}}\\
 \vdots \\
\frac{{{\partial ^\alpha }{f_{1q}}}}{{\partial {{({{(vec{\bf{X}})}^{\bf{T}}})}^\alpha }}}\\
 \vdots \\
\frac{{{\partial ^\alpha }{f_{pq}}}}{{\partial {{({{(vec{\bf{X}})}^{\bf{T}}})}^\alpha }}}
\end{array} \right] = \left[ {\begin{array}{*{20}{c}}
{\frac{{{\partial ^\alpha }{f_{11}}}}{{\partial {{({x_{11}})}^\alpha }}}}& \cdots &{\frac{{{\partial ^\alpha }{f_{11}}}}{{\partial {{({x_{m1}})}^\alpha }}}}& \cdots &{\frac{{{\partial ^\alpha }{f_{11}}}}{{\partial {{({x_{1n}})}^\alpha }}}}& \cdots &{\frac{{{\partial ^\alpha }{f_{11}}}}{{\partial {{({x_{mn}})}^\alpha }}}}\\
 \vdots &{}& \vdots &{}& \vdots &{}& \vdots \\
{\frac{{{\partial ^\alpha }{f_{p1}}}}{{\partial {{({x_{11}})}^\alpha }}}}& \cdots &{\frac{{{\partial ^\alpha }{f_{p1}}}}{{\partial {{({x_{m1}})}^\alpha }}}}& \cdots &{\frac{{{\partial ^\alpha }{f_{p1}}}}{{\partial {{({x_{1n}})}^\alpha }}}}& \cdots &{\frac{{{\partial ^\alpha }{f_{p1}}}}{{\partial {{({x_{mn}})}^\alpha }}}}\\
 \vdots &{}& \vdots &{}& \vdots &{}& \vdots \\
{\frac{{{\partial ^\alpha }{f_{1q}}}}{{\partial {{({x_{11}})}^\alpha }}}}& \cdots &{\frac{{{\partial ^\alpha }{f_{1q}}}}{{\partial {{({x_{m1}})}^\alpha }}}}& \cdots &{\frac{{{\partial ^\alpha }{f_{1q}}}}{{\partial {{({x_{1n}})}^\alpha }}}}& \cdots &{\frac{{{\partial ^\alpha }{f_{1q}}}}{{\partial {{({x_{mn}})}^\alpha }}}}\\
 \vdots &{}& \vdots &{}& \vdots &{}& \vdots \\
{\frac{{{\partial ^\alpha }{f_{pq}}}}{{\partial {{({x_{11}})}^\alpha }}}}& \cdots &{\frac{{{\partial ^\alpha }{f_{pq}}}}{{\partial {{({x_{m1}})}^\alpha }}}}& \cdots &{\frac{{{\partial ^\alpha }{f_{pq}}}}{{\partial {{({x_{1n}})}^\alpha }}}}& \cdots &{\frac{{{\partial ^\alpha }{f_{pq}}}}{{\partial {{({x_{mn}})}^\alpha }}}}
\end{array}} \right]
\end{equation}
}
In \cref{eq9}, when $\alpha = 1.0$, ${{\bf{J}}^\alpha } = {\bf{J}}$, which is the integer-order Jacobian matrix.

Now, let the matrix function ${\mathbf{F}}({\mathbf{X}}) = {{\mathbf{A}}^{p \times m}}{{\mathbf{X}}^{m \times n}} \in {\mathbb{R}^{p \times n}}$ in Definition \hyperref[def1]{1}, and the coefficient matrix ${\mathbf{A}} = \left[ {\begin{array}{*{20}{c}}
{{a_{11}}}& \cdots &{{a_{1m}}}\\
 \vdots &{}& \vdots \\
{{a_{p1}}}& \cdots &{{a_{pm}}}
\end{array}} \right]$. According to \cref{eq9}, for integer-order matrix differentiation, we have the following formula

\begin{equation}
  \label{eq10}
{\bf{J}} = \left[ {\begin{array}{*{20}{c}}
{{{\mathbf{A}}_1}}&{\mathbf{O}}& \cdots &{\mathbf{O}}\\
{\mathbf{O}}&{{{\mathbf{A}}_2}}& \cdots &{\mathbf{O}}\\
 \vdots & \vdots & \ddots & \vdots \\
{\mathbf{O}}&{\mathbf{O}}& \cdots &{{{\mathbf{A}}_n}}
\end{array}} \right]
\end{equation}
where ${\mathbf{O}}$ is a $p \times m$ zero matrix, and ${{\mathbf{A}}_1} = {{\mathbf{A}}_2} =  \cdots  = {{\mathbf{A}}_n} = {\mathbf{A}}$. And ${{\mathbf{A}}^{\mathbf{T}}} \in {\mathbb{R}^{m \times p}}$, which is the integer-order differentiation matrix. When we regard ${\mathbf{F}}({\mathbf{X}}) = {{\mathbf{A}}^{p \times m}}{{\mathbf{X}}^{m \times n}}$ as a linear layer of an ANN without bias, $\mathbf{A^T}$ can be used as an intermediate node on the computation graph and participate in the calculation of the gradient matrices of each node on the computation graph.

Similarly, the calculation formula for fractional-order matrix differentiation is as follows

\begin{equation}
  \label{eq11}
{{\bf{J}}^\alpha } = \left[ {\begin{array}{*{20}{c}}
{{{\mathbf{A}}_{11}}}&{{{\mathbf{A}}_{12}}}& \cdots &{{{\mathbf{A}}_{1n}}}\\
{{{\mathbf{A}}_{21}}}&{{{\mathbf{A}}_{22}}}& \cdots &{{{\mathbf{A}}_{2n}}}\\
 \vdots & \vdots &{}& \vdots \\
{{{\mathbf{A}}_{n1}}}&{{{\mathbf{A}}_{n2}}}& \cdots &{{{\mathbf{A}}_{nn}}}
\end{array}} \right]
\end{equation}
where $\{ {{\mathbf{A}}_{11}}, \cdots, {{\mathbf{A}}_{1n}}, \cdots, {{\mathbf{A}}_{n1}}, \cdots, {{\mathbf{A}}_{nn}}\}  \in {\mathbb{R}^{p \times m}}$. 

Since the derivative of a constant is not zero in fractional-order differentiation, in \cref{eq11}, the block matrices on the non-diagonal are not zero matrices, and the $n$ block matrices on the diagonal are also different. That is, $\mathbf{J^\alpha}$ has ${n^2}$ differentiation matrices (See Appendix \ref{appendix3}), and the transpose of each of these block matrices can be used as a differentiation matrix to participate in the chain-based derivative of fractional-order matrix differentiation. In particular, when $\alpha = 1.0$ or $n = 1$, $\mathbf{J^\alpha}$ has only one differentiation matrix, which is the integer-order differentiation matrix.

\subsection{Fractional-order Differentiation Formulas in Neural Networks}\label{sec32}

In many methods that apply the FGD method to ANNs, due to problems such as the upper and lower bounds of fractional-order differentiation, step size, polynomial calculation, and integration, they usually transform the formulas of GL, Riemann-Liouville (RL), or Caputo to meet the efficiency requirements in engineering applications. 

However, in this paper, based on the technical idea of Autograd, we want to study the fractional-order Autograd technology and use fractional-order differentiation in the neural network inside the ANN, that is, in the computation graph. Therefore, we apply fractional-order symbolic differentiation to fractional-order matrix differentiation instead of using numerical differentiations such as GL, RL, and Caputo.

Based on the discussions in previous sections and the following two reasons, we use linear functions as the discussion example.

(1) Although neural network models have become very complex and many classic models have emerged, the Linear layer based on linear functions is still an important module in many ANNs.

(2) In the experimental part of this paper, a time-series prediction task based on MLP is carried out, and its network model is composed of many Linear layers.

So, based on linear functions, we give a definition of the fractional-order differentiation formula proposed in this paper \cite{podlubny1998fractional}, as follows:

\textbf{Definition 2.} \label{def2} Let $y = xw + b$, then
\begin{equation}
  \label{eq12}
\frac{{{\partial ^\alpha }y}}{{\partial {w^\alpha }}} \coloneqq {}_0^R\mathscr{D}_w^\alpha y = \frac{x}{{\Gamma (2 - \alpha )}}{\left| w \right|^{1 - \alpha }} + sign(w)\frac{b}{{\Gamma \left( {1 - \alpha } \right)}}{\left| w \right|^{ - \alpha }}
\end{equation}
where the order $\alpha  \in (0,2)$. In this paper, to simplify the problem discussion and focus on first-order and fractional-order methods, we let $\alpha  \in (0,1]$. $sign(\bullet)$ is the sign function, which aims to ensure that the gradient direction of the corresponding term is not lost. In deep learning, $x$ is the input data, $w$ is the weight, and $b$ is the bias. See Appendix \ref{appendix4} for the derivation of \cref{eq12}.

Since the chain rule in Autograd technology is based on symbolic differentiation, \cref{eq12} as a symbolic differentiation formula meets the requirements of Autograd technology for differentiation formulas. 
At the same time, it can be noted that in \cref{eq12}, since the study of the lower bound is not the focus of this paper, its lower bound is set to 0. This is a default lower bound. In fact, the differentiation formula of Definition \hyperref[def2]{2} is obtained by statistical induction, that is, the lower bound is 0. This is consistent with the power-function differentiation formula obtained after using RL derivation. In engineering applications, different initialization methods are selected for parameters according to different models. Therefore, it should be noted that the lower bound of fractional-order differentiation here is different from the initial parameter values. At the same time, a $sign(\bullet)$ is added before the second term in \cref{eq12}. This is the same as most methods that apply fractional-order differentiation in engineering, which is to prevent complex numbers from being obtained when solving fractional powers. As a whole, this paper will also combine the gradient direction problem and explain the reason for adding the sign function in the second term from a new perspective.

Taking the ANN shown in Figure \ref{fig2} as an example, it is composed of multiple Linear layers and an MSELoss. From the scalar perspective, in integer order, the gradient direction of any leaf node is not affected by the positive or negative value of the node in the Linear layer when calculating the gradient. The gradient direction is only affected by the MSELoss, that is, the influence of the back-propagated value when the root node is differentiated. In fact, if we ignore the influence of intermediate nodes on the final gradient value, we can think that in integer order, the gradient direction is determined by the loss function. This is determined by the convex or non-convex nature of the function. In integer order, usually, we only encounter the problems of differentiating first-power and second-power functions. However, in fractional order, since the symbolic differentiation is a linear operator, there is also zero-power differentiation. Zero-power is also an even-power. So, to ensure the unity of fractional-order Autograd and integer-order Autograd, we add a sign function before zero-power differentiation. In contrast, in linear functions, no sign function is added for first-power fractional-order differentiation.

Now, let the matrix function $\mathbf{F}(\mathbf{X}) = \mathbf{A}^{p \times m}\mathbf{X}^{m \times n}+\mathbf{b}^{1 \times n}$ (In PyTorch, the bias is automatically expanded along rows to $\mathbf{b}^{p \times n}$ due to broadcasting), then according to Formulas (\ref{eq9}) and (\ref{eq11}), we can obtain any fractional-order differentiation matrix in $\frac{{\partial \mathbf{F}(\mathbf{X})}}{{\partial \mathbf{X}}}$. Taking the calculation of the first differentiation matrix $(\mathbf{A}_{11})^{\mathbf{T}}$ in \cref{eq11} as an example, where $\mathbf{A}_{11}$ can be expressed as \cref{eq13}.
\begin{equation}
  \label{eq13}
\mathbf{A}_{11} = \left[ {\begin{array}{*{20}{c}}
{\frac{{{\partial ^\alpha }{f_{11}}}}{{\partial {{({x_{11}})}^\alpha }}}}& \cdots &{\frac{{{\partial ^\alpha }{f_{11}}}}{{\partial {{({x_{m1}})}^\alpha }}}}\\
 \vdots &{}& \vdots \\
{\frac{{{\partial ^\alpha }{f_{p1}}}}{{\partial {{({x_{11}})}^\alpha }}}}& \cdots &{\frac{{{\partial ^\alpha }{f_{p1}}}}{{\partial {{({x_{m1}})}^\alpha }}}}
\end{array}} \right]
\end{equation}
It can be known that in \cref{eq14},
\begin{equation}
  \label{eq14}
\begin{array}{c}
{f_{11}} = {a_{11}}{x_{11}} + {a_{12}}{x_{21}} +  \cdots  + {a_{1m}}{x_{m1}} + {b_1}\\
{f_{21}} = {a_{21}}{x_{11}} + {a_{22}}{x_{21}} +  \cdots  + {a_{2m}}{x_{m1}} + {b_1}\\
 \vdots \\
{f_{p1}} = {a_{p1}}{x_{11}} + {a_{p2}}{x_{21}} +  \cdots  + {a_{pm}}{x_{m1}} + {b_1}
\end{array}
\end{equation}
According to \cref{eq12}, for $\left\{ {\frac{{{\partial ^\alpha }{f_{11}}}}{{\partial {{({x_{11}})}^\alpha }}}, \cdots,\frac{{{\partial ^\alpha }{f_{p1}}}}{{\partial {{({x_{11}})}^\alpha }}}, \cdots,\frac{{{\partial ^\alpha }{f_{11}}}}{{\partial {{({x_{m1}})}^\alpha }}}, \cdots,\frac{{{\partial ^\alpha }{f_{p1}}}}{{\partial {{({x_{m1}})}^\alpha }}}}\right\}$, we have \cref{eq15}.
{\footnotesize
\begin{equation}
  \label{eq15}
\left\{ \begin{array}{c}
\frac{{{\partial ^\alpha }{f_{11}}}}{{\partial {{({x_{11}})}^\alpha }}} = \frac{{{a_{11}}}}{{\Gamma (2 - \alpha )}}{\left| {{x_{11}}} \right|^{1 - \alpha }} + sign({x_{11}})\frac{{\sum\nolimits_{i = 1}^m {{a_{1i}}{x_{i1}} - {a_{11}}{x_{11}} + {b_1}} }}{{\Gamma (1 - \alpha )}}{\left| {{x_{11}}} \right|^{ - \alpha }}\\
 \vdots \\
\frac{{{\partial ^\alpha }{f_{p1}}}}{{\partial {{({x_{11}})}^\alpha }}} = \frac{{{a_{p1}}}}{{\Gamma (2 - \alpha )}}{\left| {{x_{11}}} \right|^{1 - \alpha }} + sign({x_{11}})\frac{{\sum\nolimits_{i = 1}^m {{a_{pi}}{x_{i1}} - {a_{p1}}{x_{11}} + {b_1}} }}{{\Gamma (1 - \alpha )}}{\left| {{x_{11}}} \right|^{ - \alpha }}\\
 \vdots \\
\frac{{{\partial ^\alpha }{f_{11}}}}{{\partial {{({x_{m1}})}^\alpha }}} = \frac{{{a_{11}}}}{{\Gamma (2 - \alpha )}}{\left| {{x_{m1}}} \right|^{1 - \alpha }} + sign({x_{m1}})\frac{{\sum\nolimits_{i = 1}^m {{a_{1i}}{x_{i1}} - {a_{11}}{x_{m1}} + {b_1}} }}{{\Gamma (1 - \alpha )}}{\left| {{x_{m1}}} \right|^{ - \alpha }}\\
 \vdots \\
\frac{{{\partial ^\alpha }{f_{p1}}}}{{\partial {{({x_{m1}})}^\alpha }}} = \frac{{{a_{p1}}}}{{\Gamma (2 - \alpha )}}{\left| {{x_{m1}}} \right|^{1 - \alpha }} + sign({x_{m1}})\frac{{\sum\nolimits_{i = 1}^m {{a_{pi}}{x_{i1}} - {a_{p1}}{x_{m1}} + {b_1}} }}{{\Gamma (1 - \alpha )}}{\left| {{x_{m1}}} \right|^{ - \alpha }}
\end{array} \right.
\end{equation}}

\cref{eq15} is the explicit calculation of fractional-order matrix differentiation. In computers, by converting the explicit calculation of matrix differentiation into implicit calculation, we can efficiently obtain the corresponding node's differentiation matrix and store it in the corresponding computation graph, and finally provide it to each leaf node to calculate the corresponding gradient matrix.

\subsection{Calculation of Fractional-Order Gradient Matrix}\label{sec33}

The core problem of this paper is to use $\mathbf{J}^\alpha$ to realize the application of the FGD method in hidden layers. As described in Section \ref{sec22}, Case 1, 2, and 3 are essentially the same type of problem. Therefore, to simplify the problem, the algorithms and experiments in this paper only take Case 1 as an example. And the ANN in the experimental part is also based on the network model of \cref{eq1} to conduct qualitative and quantitative analyses on real-world time-series datasets.

As can be seen from \cref{eq15}, in Section \ref{sec22}, fractional-order matrix differentiation is obtained through explicit calculation. However, in application scenarios, this is very time-consuming. Therefore, in this section, we first need to convert the fractional-order matrix differentiation in Section \ref{sec22} into implicit calculation, that is, all calculations are based on matrices.

According to \cref{eq1} and Figure \ref{fig2}, in the neural network model example in this paper, the tensor form of any Linear layer can be expressed as \cref{eq16}. Note that in PyTorch, the bias is automatically expanded along rowa to $\mathbf{b}^{p \times n}$ due to broadcasting.
\begin{equation}
  \label{eq16}
\mathbf{Y}^{p \times n} = \mathbf{X}^{p \times m}\mathbf{W}^{m \times n}+\mathbf{b}^{1 \times n}
\end{equation}

When solving the gradients of each node in \cref{eq16}, let the required back-propagation matrix be $\mathbf{G}$, $\mathbf{G}=\left[ {\begin{array}{*{20}{c}}
{{g_{11}}}& \cdots &{{g_{1n}}}\\
 \vdots &{}& \vdots \\
{{g_{p1}}}& \cdots &{{g_{pn}}}
\end{array}} \right] \in \mathbb{R}^{p \times n}$, then, according to Case 1, the gradient matrices of the weight matrix $\mathbf{W}$, the intermediate node $\mathbf{X}$, and the bias $\mathbf{b}$ can be expressed as \cref{eq17}.
\begin{equation}
  \label{eq17}
\begin{array}{l}
\nabla _{\mathbf{W}}^\alpha {\mathbf{Y}} = {\left( {{{\mathbf{X}}_{ij}}} \right)^{\mathbf{T}}} \bullet {\mathbf{G}}\\
{\nabla _{\mathbf{X}}}{\mathbf{Y}} = {\mathbf{G}} \bullet {{\mathbf{W}}^{\mathbf{T}}}\\
{\nabla _{\mathbf{b}}}{\mathbf{Y}} = \sum\nolimits_{k = 1}^p {{{\mathbf{G}}_{kn}}} 
\end{array}
\end{equation}
where $\bullet$ represents matrix multiplication. $\left( {{{\mathbf{X}}_{ij}}} \right)^{\mathbf{T}}$ represents the fractional-order differentiation matrix for the $(i,j)$-th block $\mathbf{J}^\alpha$. $\sum\nolimits_{k = 1}^p {\mathbf{G}_{kn}}$ represents summing the elements of each column for $\mathbf{G}$ to obtain a row vector of size $1 \times n$, that is, the gradient of $\mathbf{b}$. In the Linear layer, which is also a leaf node, the weight matrix usually has a much greater impact on the output result of the ANN than the bias. Therefore, for the convenience of the qualitative discussion in the experimental part of Section \ref{sec4}, in \cref{eq17}, only the gradient matrix of $\mathbf{W}$ is obtained through fractional-order matrix differentiation, and the gradient matrix of $\mathbf{b}$ is obtained through integer-order matrix differentiation.

Now, if we take the first block matrix in the fractional-order Jacobian matrix as the differentiation matrix. In \cref{eq17}, it is $\mathbf{X}_{11}$. Then, the implicit calculation process of $\nabla _{\mathbf{W}}^\alpha \mathbf{Y}$ can be shown as Algorithm \ref{alg1}.
\begin{algorithm*}[h]
    \caption{Calculations of the weight matrix gradient based on $\mathbf{X}_{11}$} 
	\label{alg1} 
	\begin{algorithmic}[1]
        \REQUIRE $\mathbf{W} \in \mathbb{R}^{m \times n},\mathbf{X} \in \mathbb{R}^{p \times m},\mathbf{b} \in \mathbb{R}^{1 \times n},\mathbf{G} \in \mathbb{R}^{p \times n},\alpha$
        \ENSURE $\mathbf{W_{grad}}$ \hfill \scalebox{1.2}[1.5]{$\triangleright$} $\mathbf{W_{grad}}$ denotes $\nabla _{\mathbf{W}}^\alpha \mathbf{Y}$.
        \STATE $\mathbf{F = W}[:,0].view(1, - 1) \in \mathbb{R}^{1 \times m}$ ~~~~\hfill \scalebox{1.2}[1.5]{$\triangleright$} In order to obtain $\mathbf{A_{11}}$, the first column of $\mathbf{W}$ is extracted and then transposed.
        \STATE Calculate the first part of $\mathbf{W}$, which is the first term in \cref{eq12}:
        \STATE $\mathbf{W_{main}} = \mathbf{X} \odot \left|\mathbf{F}\right|^{\frac{{1 - \alpha }}{{\Gamma(2 - \alpha )}}} \in \mathbb{R}^{p \times m}$
        \STATE Calculate the second part of $\mathbf{W}$, which is the second term in \cref{eq12}. The calculation involves $\mathbf{W_{partial1}}$, $\mathbf{W_{partial2}}$, $\mathbf{W_{partial3}}$, and $\mathbf{W_{partial4}}$. Refer to \cref{eq15} for details:
        \STATE $\mathbf{W_{partial1}} = (\mathbf{X} \bullet {{\mathbf{F}}^{\mathbf{T}}}).view( - 1,1) \in \mathbb{R}^{p \times 1}$
        \STATE $\mathbf{W_{partial2}} = (\mathbf{W_{partial1}}).expand( - 1,m) \in \mathbb{R}^{p \times m}$
        \STATE $\mathbf{bias} = torch.full((p,m),{\mathbf{b}}[0].item()) \in \mathbb{R}^{p \times m}$ \hfill \scalebox{1.2}[1.5]{$\triangleright$} Specify ${\mathbf{b}}[0]$ as a matrix of size $p \times m$ 
        \STATE $\mathbf{W_{partial3}}{\mathbf{ = }}({\mathbf{W_{partial2}}} - {\mathbf{X}} \odot {\mathbf{F + bias}}) \in \mathbb{R}^{p \times m}$
        \STATE $\mathbf{W_{partial4}}{\mathbf{ = }}sign({\mathbf{F}}) \odot {\left| {\mathbf{F}} \right|^{\frac{{ - \alpha }}{{gamma(1 - \alpha )}}}} \in \mathbb{R}^{p \times m}$
        \STATE $\mathbf{W_{partial}} = {\mathbf{W_{partial3}}} \odot {\mathbf{W_{partial4}}} \in \mathbb{R}^{p \times m}$
        \hfill \scalebox{1.2}[1.5]{$\triangleright$} Obtain the second part of $\mathbf{W}$. See \cref{eq12,eq15}.
        \STATE $\mathbf{W_{differentiation}} = {(\mathbf{W_{main}}} + \mathbf{W_{partial}})^{\mathbf{T}} \in \mathbb{R}^{m \times p}$ 
        \hfill \scalebox{1.2}[1.5]{$\triangleright$} Obtain the differentiation of $\mathbf{W}$.
        \STATE $\mathbf{W_{grad} = W_{differentiation}} \bullet \mathbf{G} \in \mathbb{R}^{m \times n}$ 
        \hfill \scalebox{1.2}[1.5]{$\triangleright$} Obtain the gradient of $\mathbf{W}$.
	\end{algorithmic}
\end{algorithm*}

In Algorithm \ref{alg1}, it mainly calculates the first and second terms of \cref{eq12}. According to \cref{eq12}, in tensor form, the first term is actually a linear mapping in integer order, and its calculation is relatively simple, corresponding to line 3 in Algorithm \ref{alg1}. In Case 4, many fractional-order differentiations on loss functions are also linear mappings in integer order, and they are equivalent to only calculating line 3. From line 4 to 10, it is the calculation of the second term of \cref{eq12}. We define the obtained differentiation matrix as the fractional-order term differentiation matrix. Although the calculation of the fractional-order term differentiation matrix is relatively complex, it ensures the essential difference between the final obtained differentiation matrix and that in integer order. That is, it cannot be obtained by simply fine-tuning the learning rate or using a special learning rate scheduling strategy.

In Algorithm \ref{alg1}, 10 steps of operations are carried out to obtain the gradient of the weight matrix $\mathbf{W}$. In contrast, in integer-order differentiation, ${\nabla _{\mathbf{W}}}{\mathbf{Y = }}{{\mathbf{X}}^{\mathbf{T}}} \bullet {\mathbf{G}}$. Compared with integer-order matrix differentiation, the method in this paper increases in computational complexity. However, since both are based on implicit calculation, this increase is linear and controllable. And when $\alpha = 1.0$, using Algorithm \ref{alg1} can get $\nabla _{\mathbf{W}}^1{\mathbf{Y = }}{{\mathbf{X}}^{\mathbf{T}}} \bullet {\mathbf{G}}$, which is the same as the result obtained by integer-order matrix differentiation.

Algorithm \ref{alg1} finally obtains the gradient of the leaf node. When it replaces the gradients in the GD method and its variant methods and participates in the iterative update of the parameter matrix, it will form the corresponding fractional-order optimization method, such as the fractional-order SGD (FSGD) optimizer and the fractional-order Adam (FAdam) optimizer. Here, taking the FSGD optimizer defined in this paper as an example, we analyze its time complexity and convergence.

\textbf{Time Complexity Analysis:} Although the FSGD optimizer needs to carry out multiple steps when solving the fractional-order gradient matrix and involves a differentiation matrix with a linear mapping to integer-order and a fractional-order term differentiation matrix. However, like the SGD and Adam optimizers, their gradient matrices are based on implicit calculation, that is, the time complexity of the FSGD optimizer is the same as that of its corresponding integer-order optimization method. Here, assuming that given the network model and the $batch$ of training samples, we can set the time complexity of training one $batch$ as $O(d)$. If the total number of iterations $epoch$ is $n$, then the total time complexity of the FSGD optimizer is $O(n*d)$, which is the same as that of the SGD, Adam, and FAdam optimizers.

\textbf{Convergence Analysis:} As shown in Figure \ref{fig1}, existing fractional-order methods improve in Optimizer.step(), which is an operation on the integer-order gradient matrix of the computation graph. When the operation involves a non-linear transformation of the matrix, their convergence characteristics will be different, and separate convergence analysis is required. On the contrary, the method proposed in this paper is an operation to solve the fractional-order gradient matrix of the leaf node in Loss.backward(). When we assume that the objective function boundary is bounded and the fractional-order gradient is bounded, their convergence characteristics are the same as those of their corresponding integer-order optimizers. The change of the fractional order has a linear impact on the convergence speed of $Loss$ \cite{HARJULE2025116009}. Therefore, there is no need to list separate formulas for proof. That is, when given the same hyperparameters, the convergence speeds of the two are the same. In many FGD methods applied to ANN training, although they have a faster convergence speed than integer order, this difference in convergence speed is actually equivalent to the impact of different learning rates on the convergence trajectory in integer-order optimizers. In addition, in engineering applications, since training often uses an early-stopping mechanism and real-world datasets are usually biased. Therefore, the global optimum on the training set is not necessarily the optimum on the test set. In conclusion, this paper pays more attention to the performance of the network model after using FGD method training. Therefore, the relevant discussion is further carried out in the experimental part of Section \ref{sec4}.

\section{Experiment}\label{sec4}

This section designs a simple ANN training task and verifies the method on two types of real-world datasets. It conducts quantitative and qualitative analyses on the performance of the method.

\subsection{Overall Structure of the Neural Network}\label{sec41}

\begin{figure*}[h]
  \centering
  \includegraphics[width=\textwidth]{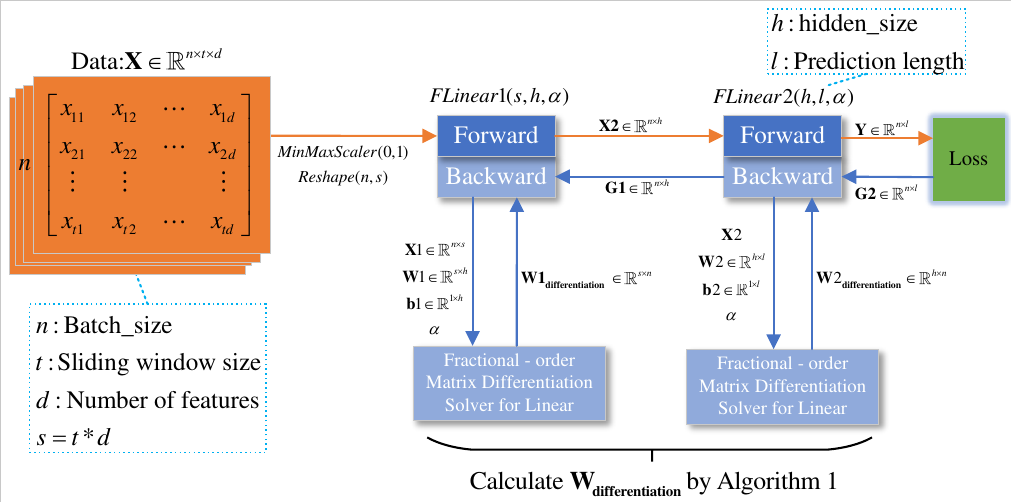}
  \caption{The overall structure of the paper.}
  \label{fig3}
\end{figure*}

The neural network used in the experiment is shown in \cref{eq1} and the computation graph in Figure \ref{fig2}. When FLinear is used to replace Linear, its overall structure during training can be summarized in Figure \ref{fig3}.

Figure \ref{fig3} is not a SOTA model. Its purpose is to facilitate the quantitative and qualitative analyses of the FGD method in this paper. It has two Linear layers and involves the problem of fractional-order matrix differentiation in the back-propagation of ANNs, meeting the requirements of the research on ANNs.

In Figure \ref{fig3}, FLinear only adds an incoming parameter $\alpha$. In Forward, FLinear is the same as Linear. Their differences lie in Backward. Compared with Linear, FLinear's Backward has a separately designed Fractional-order Matrix differentiation Solver. The role of the solver is to calculate the fractional-order differentiation matrices of each node, and the specific calculation method corresponds to Algorithm \ref{alg1}. As described in the previous Section \ref{sec3}, the solver is the specific engineering implementation of the method in this paper.

Just like in integer-order Autograd, in Backward, the fractional-order differentiation matrix is multiplied by the back-propagated matrix to obtain the gradient matrix of the node. Then, the gradient matrices of each leaf node are saved in the computation graph for the corresponding optimizer to call and participate in the iterative update of the parameter matrix.

\subsection{Preparation for the Experiment}\label{sec42}

\textbf{Experimental Environment Configuration:} Since Section \ref{sec43} involves the analysis of memory size and time consumption, the specific experimental environment configuration is introduced here. The CPU is Intel(R) Xeon(R) CPU E5-2699 v4 $\symbol{64}$ 2.20GHz. The GPU is NVIDIA GeForce RTX 2080 Ti. The IDE is Jupyter 7.2.2. The programming language is Python 3.12.7. The deep-learning framework is PyTorch 2.6.0 $+$ CUDA 12.6.

\textbf{Datasets and Preprocessing:} In the experiment, two real-world time-series prediction datasets are used, namely Dow Jones Industrial Average (DJI) and the public dataset ETTh1. DJI is a dataset with spikes, strong fluctuations, and strong noise, while ETTh1 is the opposite. This is to analyze the performance of the FGD method on different types of datasets. Their detailed information is shown in Table \ref{tab1}.

\begin{table*}[h]
  \centering
  {\footnotesize
  \caption{The statistics of ETTh1 and DJI}
  \label{tab1}
  \begin{tabular*}{\textwidth}{@{\extracolsep{\fill}}ccccccc@{}}
  %\begin{tabular}{c|c|c|c|c|c|c}
    \hline
    Datasets & Features & Timesteps & Granularity & Timespan & \makecell{Training: \\ Validation: \\ Test} & Labels \\ \hline
    DJI & 5 & 5,969 & 1day & \makecell{Jan 3, 2000 - \\ Sep 22, 2023} & 7:2:1 & Close \\ 
    ETTh1 & 7 & 17,420 & 1hour & \makecell{Jul 1, 2016 - \\ Jun 26, 2018} & 7:2:1 & OT \\
    \hline
  \end{tabular*}}
\end{table*}

In Table \ref{tab1}, the five features of DJI are Open, High, Low, Volume, and Close. The Close value is predicted jointly by the five feature dimensions. The seven features of ETTh1 are HUFL, HULL, MUFL, MULL, LUFL, LULL, and OT. The OT value is predicted jointly by the seven feature dimensions. There are no outliers in either dataset, so data cleaning is not required. As shown in Figure \ref{fig3}, for data preprocessing, only $Min - Max$ normalization and simple mini-batching are carried out.

\textbf{Hyperparameter Settings:} To fairly show the trajectories of $loss$ values and quantitative metric values of different orders, some hyperparameter selections and settings are made in the experiment:

\begin{enumerate}[label=(\arabic*)]
    \item Set random seeds for batching and weight matrix initialization. When iterating at different orders, ensure that the shuffling results of batching can be reproduced and the initialized weight matrices are the same. This can eliminate the influence of randomness on the results.
    \item During the experiment, set the learning rate uniformly as $lr = 1e - 4$. In many experiments on FGD method research, the influence of the learning rate on the convergence trajectory is obvious. Especially for some FGD methods, the obtained gradient matrix may actually be equivalent to providing a scalar outside the integer-order gradient matrix. This scalar changes the magnitude of the learning rate. That is, due to the change of order, the integer-order gradient matrix gets a larger or smaller learning rate for scalar matrix multiplication. Therefore, setting an appropriate learning rate can ensure that the performance display of the FGD method is not affected by the learning rate.
    \item \label{setting1}As shown in Figure \ref{fig1}, the method in this paper is to solve the fractional-order gradient matrices of each leaf node on the computation graph. During parameter updating, the optimizer is selected according to the task requirements. In the whole experiment process of this paper, the SGD optimizer is selected, that is, the FSGD optimizer. Compared with other optimizers (such as the popular Adam), the SGD optimizer will not change the characteristics of the parameter matrix. Other optimizers will perform various non-linear operations on the gradient matrix after obtaining the leaf node gradient matrix, destroying the additivity and multiplicativity of the gradient matrix, thus changing the characteristics of the fractional-order gradient matrix. Finally, it will affect our qualitative and quantitative analysis of the FGD method. Similarly, for the parameter settings of the SGD optimizer, we set $weight\_decay = 0$ and $momentum = 0$. Because these two hyperparameters will also destroy the additivity of the fractional-order gradient matrix.
    \item Similarly, the ANN used in the experiment does not set an activation function. Because the activation function will also destroy the additivity and multiplicativity of the fractional-order gradient matrix, affecting our qualitative and quantitative analysis of the results.
    \item According to Figure \ref{fig3}, set these hyperparameters uniformly: the sliding window size is 36, $Batch\_size = 256$, the prediction length is 48, and the loss function uses MSELoss. According to the characteristics of DJI and ETTh1, to better show the convergence trajectories of different orders, when training DJI, set $hidden\_size = 256$. When training ETTh1, set $hidden\_size = 128$.
\end{enumerate}

\subsection{Experimental Results Performance Analysis} \label{sec43}

\begin{figure*}[h]
  \centering
  \includegraphics[width=\hsize]{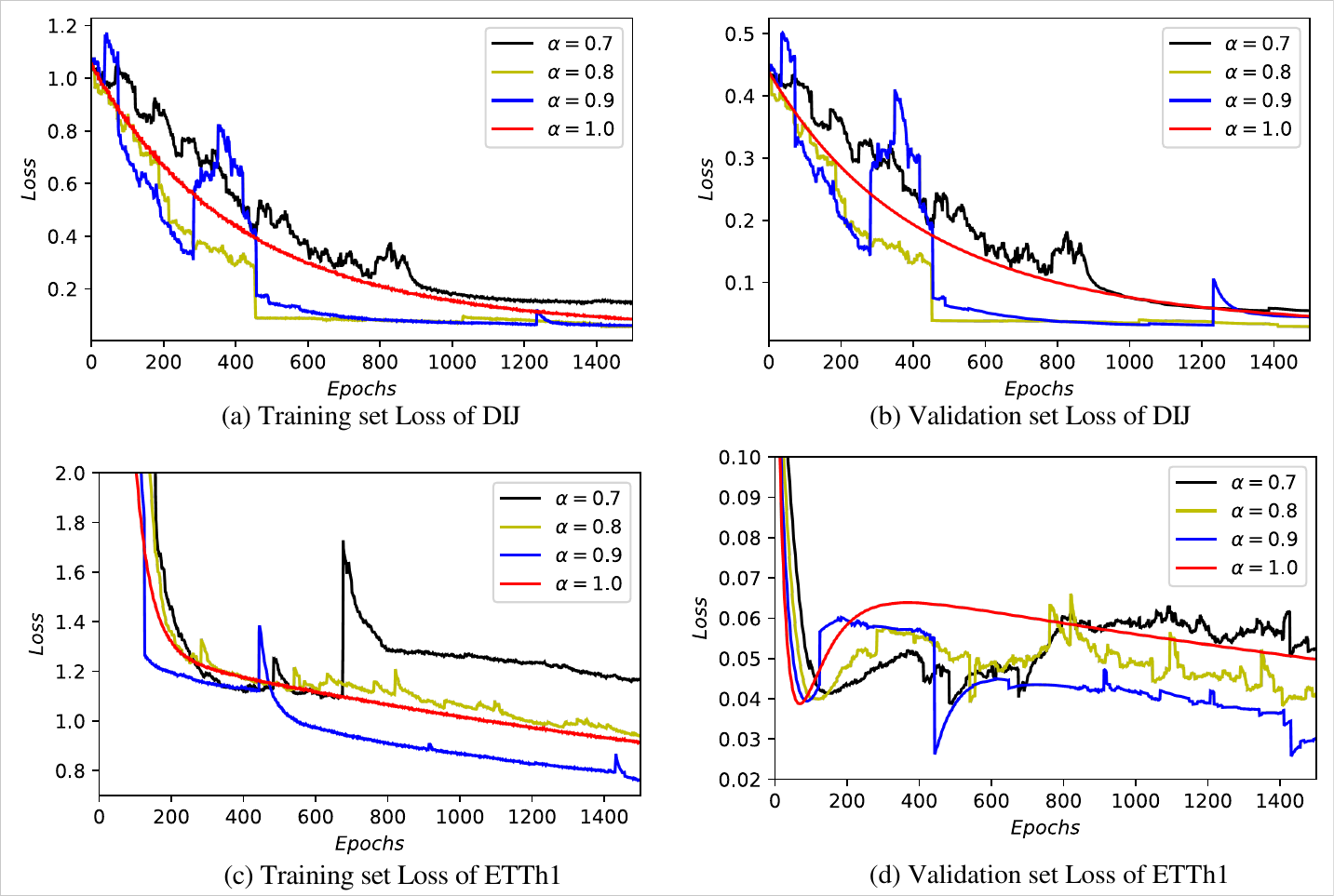}
  \caption{The convergence curves of DJI and ETTh1 on the training and validation sets.}
  \label{fig4}
\end{figure*}

We respectively train the training sets of two datasets and display their convergence trajectories at 1500 iterations. Meanwhile, after each iteration, the model is used for quantitative evaluation on the validation set, with the evaluation metric being MSE, i.e., MSELoss. Experiments with different orders are carried out, and the results of orders $\alpha = \{0.7, 0.8, 0.9, 1.0\}$ are collected and shown in Figure \ref{fig4}.

Overall, Figure \ref{fig4} shows the characteristics of the FSGD optimizer that are different from those of the SGD optimizer, including the self-contained learning rate scheduling strategy and regularization. The details are as follows:

\begin{enumerate}[label=(\arabic*)]
    \item Although there are various interfering factors, it can still be seen that, at most fractional orders, due to the influence of the learning rate scheduling strategy of the FSGD optimizer, the $Loss$ of the two datasets often decreases faster than that of the integer order in the early stage of iterations. Of course, this difference in convergence speed is at a linear level and has not reached the change level from the SGD optimizer to the Adam optimizer.
    \item It can also be seen in Figure \ref{fig4}(d) that, during the fractional-order optimization process, due to the special regularization characteristic of the FSGD optimizer, even in the case of overfitting, they can jump out of the local optimal solution and search for the global optimal solution. In contrast, once the integer order overfits and falls into a local optimal solution, it cannot jump out.
    \item Overall, compared with the SGD optimizer, the convergence trajectory of the FSGD optimizer is not smooth. This is caused by the fractional-order term in \cref{eq12}. When performing fractional-order matrix differentiation, even though this method only extracts a block matrix from the Jacobian matrix differentiation, the fractional-order differentiation matrix of the weight matrix is affected by the input matrix and the bias vector. In ANNs, because of the early-stopping mechanism, we usually only save the model with the smallest $Loss$ on the validation set. This non-smooth characteristic on the convergence trajectory will not have a great impact on performance.
    \item Since we finally obtain the model at the moment when the $Loss$ on the validation set is the smallest, by comparing the convergence trajectories of different orders in Figure \ref{fig4}(b) and (d), it can be found that the performance of $\alpha = 0.7$ is generally worse than that of the integer order, and $\alpha = 0.8$ is the second. By analyzing the fractional-order differentiation term in \cref{eq12}, it can be seen that when the value of $\alpha$ approaches 0 more and more, the value of the differentiation will become more and more extreme. From the perspective of the FSGD optimizer, it means gradient explosion. The training of the ANN model has almost no opportunity to give full play to the unique advantages of the FSGD optimizer and will become worse and worse due to gradient explosion. This also verifies the results of some previous studies \cite{wang2022study,zhou2025fractional,zhou2025improved}, that is, the optimal fractional-order selection of the FGD method is usually $0.9 < \alpha < 1.0$.
\end{enumerate}

\begin{figure*}[h]
  \centering
  \includegraphics[width=\hsize]{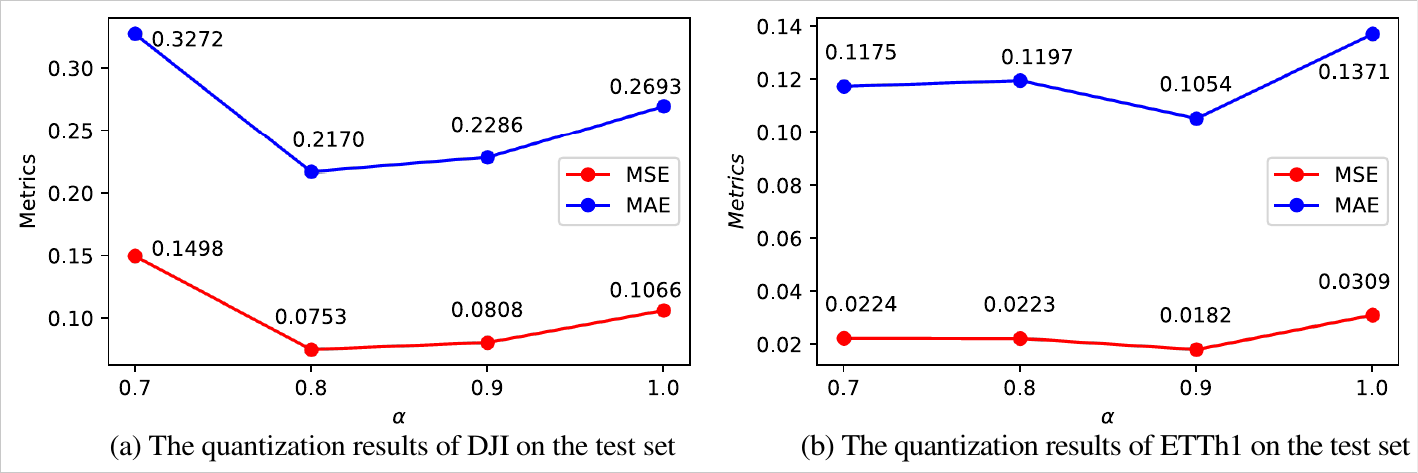}
  \caption{The test set metrics on different $\alpha $.}
  \label{fig5}
\end{figure*}

In Figure \ref{fig4}(b) and (d), the convergence trajectory of $Loss$ can already well show the performance of each order. However, for quantitative analysis, we use the optimal models to make quantitative evaluations on the test set. In addition to MSE, the mean absolute error (MAE) is also used as the evaluation metric. They are commonly used evaluation metrics in regression tasks, and a smaller value indicates better performance. The detailed quantitative results are shown in Figure \ref{fig5}.

Overall in Figure \ref{fig5}, the performance of the fractional-order orders $\alpha = \{0.8, 0.9\}$ is better. This verifies the results and speculations of the validation set. In addition, in Figure \ref{fig5}(a), the order change is more sensitive to the results. As mentioned before, DJI has strong noise. This means that it needs more regularization to prevent overfitting. The FSGD optimizer has self-contained regularization and will work better for DJI-type datasets. Therefore, after selecting an appropriate order, the performance improvement will be obvious. Similarly, the results in Figure \ref{fig5}(b) are reverse verification. Since ETTh1 is a stationary dataset with little noise, the change in the quantized results is not as large as that of DJI when the order changes. Finally, we can see that in DJI, the results of $\alpha = 0.7$ deteriorate rapidly. It is the only fractional order with worse performance than the integer order. This verifies the analysis in the fourth point of Figure \ref{fig4}.

\textbf{Analysis of Time Consumption and GPU Memory Usage:} To analyze the time consumption and GPU memory usage of the FSGD optimizer (in this paper, it refers to SGD $+$ FLinear) and the integer-order optimizer, we record the running time and GPU memory at different orders for the first epoch of ETTh1's training set. Their detailed information is shown in Table \ref{tab2}.
\begin{table*}[h]
  \centering
  {\footnotesize
  \caption{The comparison of time consumption and GPU Memory usage at different orders.}
  \label{tab2}
  \begin{tabular*}{\textwidth}{@{\extracolsep{\fill}}cccc@{}}
    \hline
    Method & $\alpha$ & Time (s) & Memory (MB) \\
    \hline
    FSGD & 0.9 & 0.7821 & 38.4521 \\
    FSGD & 1.0 & 0.5873 & 38.4521  \\
    SGD & $\#$ & 0.3041 & 38.4521 \\
    \hline
  \end{tabular*}}
\end{table*}

In Table \ref{tab2}, it can be found that whether directly using the SGD optimizer, and using integer orders or fractional orders in the FSGD optimizer, their GPU memory usage is the same. As known from Figure \ref{fig1} and \ref{fig3}, the FSGD optimizer (including the FAdam optimizer) actually changes the calculation method of the gradient matrix of each node on the computation graph. Given the model and data, the size of the computation graph is fixed and will not increase the GPU memory due to the addition of a fractional-order matrix differentiation solver in $Loss.Backward(retain\_graph=True)$.

In Table \ref{tab2}, we focus on analyzing the time consumption under three conditions. First, although the time consumptions of the three are different, they still change at a linear level. This verifies the time-complexity analysis of Algorithm \ref{alg1}. Second, the FSGD optimizer takes more time than the SGD optimizer. This is because calculating the fractional-order matrix differentiation requires more intermediate processes, and a certain amount of time needs to be consumed in these processes. Third, in the FSGD optimizer, the time consumption of $\alpha = 0.9$ is more than that of $\alpha = 1.0$. Through the discussion of the previous \cref{eq12} and Algorithm \ref{alg1}, it can be known that the training result of the FSGD optimizer with $\alpha = 1.0$ is equal to that of the SGD optimizer, but compared with $\alpha = 0.9$, $\alpha = 1.0$ can eliminate the calculation of the fractional-order differentiation term, thus saving most of the calculations in the fractional-order matrix differentiation solver and reducing time consumption.

Through the display and analysis of Figures \ref{fig4} and \ref{fig5} and Table \ref{tab2}, we can verify the performance of the FSGD optimizer, and then verify the superiority of the fractional-order matrix differentiation theory, as well as the feasibility of ${{\bf{J}}^\alpha }$ in engineering applications.

\section{Summary} \label{sec5}

This paper studies the fractional-order matrix differentiation method in ANNs. By calculating the fractional-order differentiation matrix using the ${{\bf{J}}^\alpha }$ method, it proposes a novel solution for the research of the FGD method, solving the problem that the FGD method cannot be applied inside ANNs. The feasibility of the method is verified from both theoretical and experimental perspectives, laying a foundation for further research and improvement of the FAutograd technology. In practical engineering applications, by replacing Linear with FLinear, the optimizer becomes a fractional-order optimizer. For example, in the experiments, when the optimizer during training is changed from SGD to Adam, the FSGD optimizer becomes the FAdam optimizer. However, as described in the hyperparameter setting (\hyperref[setting1]{3}), adaptive learning rate optimizers will change the original characteristics of the fractional-order gradient matrix, affecting our qualitative and quantitative research on ${{\bf{J}}^\alpha }$. Therefore, the FAdam optimizer is not selected in the experimental part.

In future work, we will further study and improve FLinear. Moreover, we will reconstruct modules such as CNN, RNN, and Transformer into their corresponding fractional-order modules, namely FCNN, FRNN, and FTransformer. Finally, we will complete the construction of the FAutograd framework to enable the application of fractional-order matrix differentiation in more types of ANNs.

From the discussion in the experimental part, we can find that in practical engineering applications, the FGD method is more prone to gradient explosion than the GD method. This situation is more obvious when the fractional order is extreme settings. In some cases, the FGD method will experience performance degradation due to the gradient explosion problem. We know that this is caused by the fractional-order differentiation term. Therefore, in the practical engineering applications of ANNs, we will start from the fractional-order symbolic differentiation formula and further optimize it. This is our future research direction.

\begin{appendices}
\section{Convergence Trajectories and Gradient Directions} \label{appendix1}
\setcounter{figure}{0}
\renewcommand{\thefigure}{A.\arabic{figure}}
\setcounter{equation}{0}  
\renewcommand{\theequation}{A.\arabic{equation}} 

Let the scalar function $y=(x_1 + 2)^2+(x_2 + 3)^2$. It can be known that the extreme point of the function is $(x_1,x_2)=(-2,-3)$. According to \cref{eq12}, the update formulas of the two parameters are as follows:
{\footnotesize
\begin{equation}
  \label{eqA1}
    x_1^{(k + 1)} = x_1^{(k)} - \eta \bullet \text{sign}(x_1^{(k)} + 2)\left(\frac{2}{\Gamma(3 - \alpha)}{\left| {x_1^{(k)} + 2} \right|^{2 - \alpha }}+\frac{{(x_2^{(k)} + 3)^2}}{\Gamma(1 - \alpha)}{\left| {x_1^{(k)} + 2} \right|^{ - \alpha }}\right)
\end{equation}}
{\scriptsize
\begin{equation}
  \label{eqA2}
  \small
    x_2^{(k + 1)} = x_2^{(k)} - \eta \bullet \text{sign}(x_2^{(k)} + 3)\left(\frac{{(x_1^{(k)} + 2)^2}}{\Gamma(1 - \alpha)}{\left| {x_2^{(k)} + 3} \right|^{ - \alpha }}+\frac{2}{\Gamma(3 - \alpha)}{\left| {x_2^{(k)} + 3} \right|^{2 - \alpha }}\right)
\end{equation}}
where $\eta$ is the learning rate and $k = 1,2,...,N$.

\begin{figure*}[h]
    \centering
    \label{figA1}
    \includegraphics[width=0.6\hsize]{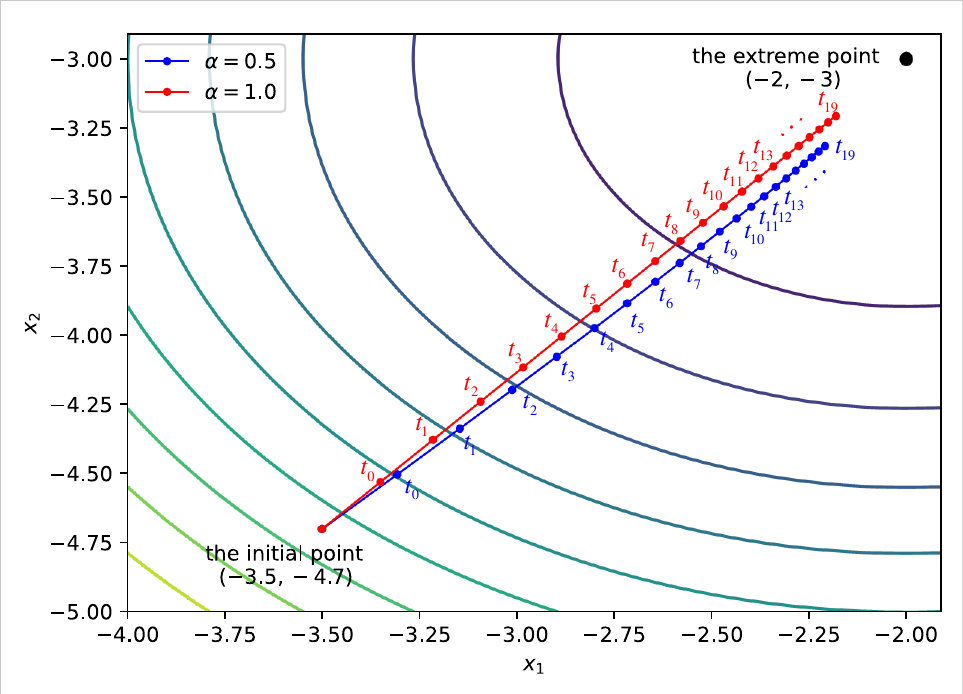}
    \caption{The convergence trajectories of the parameters in the fractional order and the integer order.}
\end{figure*}

From Formulas (\ref{eqA1}) and (\ref{eqA2}), the convergence trajectories of the parameters during iterations can be obtained. When $\alpha = \{0.5,1.0\}$, we randomly set $(x_1^0,x_2^0)=(-3.5,-4.7)$ (which are also the lower bounds of their respective fractional-order differentiations) and record the results of the first 20 iterations. See Figure \hyperref[figA1]{A.1}.

In Figure \hyperref[figA1]{A.1}, it can be seen that compared with integer-order differentiation, the convergence trajectory of fractional-order differentiation shows the characteristic of being fast at first and then slow. Usually, in deep-learning optimization methods, such an effect can only be achieved by calling a learning rate scheduler. Therefore, fractional-order differentiation is actually equivalent to having a special learning rate scheduling strategy.

Regarding the characteristics of fractional-order at saddle points, given a ternary function $z = x^2 - y^2$, draw its three-dimensional graph. When the parameters are at the saddle point, we solve the gradients respectively using integer-order differentiation and fractional-order differentiation, that is, using the GD method and the FGD method respectively. See Figure \hyperref[appendix1]{A.2}.

\begin{figure*}[h]
    \centering
    \label{figA2}
    \includegraphics[width=0.8\hsize]{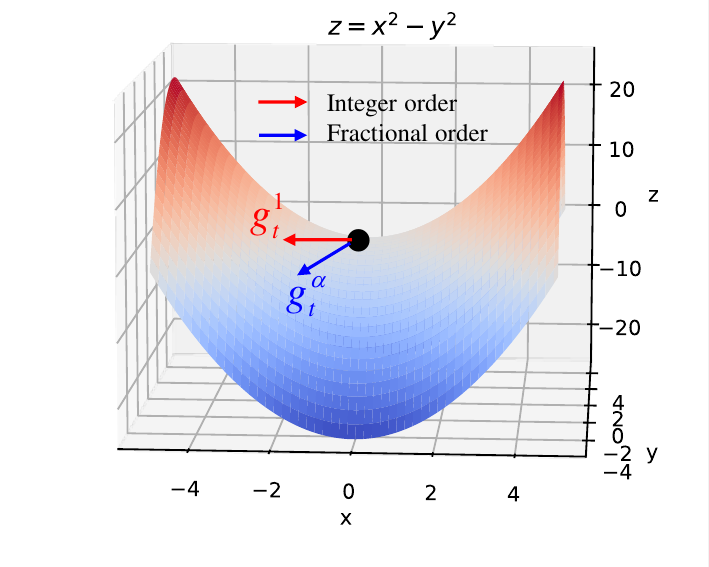}
    \caption{The gradient directions at the saddle points for the fractional order and the integer order.}
\end{figure*} 

As is well known, integer-order differentiation will cause the parameters to oscillate left and right at the saddle point. On the contrary, according to \cref{eq12} and Figure \hyperref[appendix1]{A.2}, due to the influence of the fractional-order differentiation term, when $\alpha\neq1.0$, the gradient vector must not be parallel to the $xz$ plane. Therefore, the FGD method will not get stuck at the saddle point.

\section{Proof of Fractional-order Differentiation Regularization in Artificial Neural Networks} \label{appendix2}
\renewcommand{\thefigure}{B.\arabic{figure}}
\setcounter{figure}{0}
\renewcommand{\theequation}{B.\arabic{equation}} 
\setcounter{equation}{0} 
\begin{proof}
Let $J(w)$ be the objective function. For the GD method, we have a regularization method with a penalty function. That is,

\begin{equation}
    \label{eqB1}
    \tilde J(w) = J(w) + p(w)
\end{equation}
where $\tilde J(w)$ is the total objective function and $p(w)$ is the penalty function. Specifically, ${L^1}$ and ${L^2}$ are penalty functions when $p(w)$ takes specific forms.

When using integer order to calculate the partial derivative of $\tilde J(w)$, we have
{\footnotesize
\begin{equation}
    \label{eqB2}
    \begin{aligned}
    \tilde J'(w) & = J'(w) + p'(w) = J'(y)y'(x) + p'(w)
    \end{aligned}
\end{equation}}

In contrast, for an objective function without a penalty function, using fractional-order differentiation and considering $y = wx + b$, we have
{\small
\begin{equation}
    \label{eqB3}
    \begin{aligned}
    J'(w) &  = J'(y)\frac{{\partial ^\alpha y}}{{\partial {w^\alpha }}} = J'(y) (\frac{x}{{\Gamma (2 - \alpha )}}{\left| w \right|^{1 - \alpha }} + \text{sign}(w)\frac{b}{{\Gamma \left( {1 - \alpha } \right)}}{\left| w \right|^{ - \alpha }})\\
     &= J'(y)\frac{x}{{\Gamma (2 - \alpha )}}{\left| w \right|^{1 - \alpha }} + J'(y) \text{sign}(w)\frac{b}{{\Gamma \left( {1 - \alpha } \right)}}{\left| w \right|^{ - \alpha }}
    \end{aligned}
\end{equation}}

In \cref{eqB3}, let ${J'_1}(w) = J'(y)\frac{x}{{\Gamma (2 - \alpha )}}{\left| w \right|^{1 - \alpha }}$ and ${p'_1}(w) = J'(y) \text{sign}(w)\frac{b}{{\Gamma \left( {1 - \alpha } \right)}}{\left| w \right|^{ - \alpha }}$. Then, according to \cref{eqB2}, we have

\begin{equation}
    \label{eqB4}
    J'(w)  = {J'_1}(w) + {p'_1}(w)
\end{equation}

From \cref{eqB4}, it can be seen that when using fractional-order differentiation, the objective function is equivalent to having a built-in penalty function. Therefore, fractional-order differentiation has the characteristic of being equivalent to having built-in regularization. Moreover, this kind of regularization is not a linear mapping with ${L^1}$ and ${L^2}$.
\end{proof}

\section{Proof of the Number of Fractional-order Jacobian Matrix Differentiation Block Matrices} \label{appendix3}
\renewcommand{\thefigure}{C.\arabic{figure}}
\setcounter{figure}{0}
\renewcommand{\theequation}{C.\arabic{equation}} 
\setcounter{equation}{0} 
\begin{proof}
According to \cref{eq8}, let $\mathbf{F}(\mathbf{X})=\mathbf{A}^{p\times m}\mathbf{X}^{m\times n}\in\mathbb{R}^{p\times n}$, then

\begin{equation}
    \label{eqC1}
    \mathbf{J}^{\alpha}\in\mathbb{R}^{pn\times mn}
\end{equation}
Since $\mathbf{A}\in\mathbb{R}^{p\times m}$, the integer-order differentiation matrix of $\mathbf{F}(\mathbf{X})$ w.r.t. $\mathbf{X}$ is

\begin{equation}
    \label{eqC2}
    \mathbf{A}^T\in\mathbb{R}^{m\times p}
\end{equation}

To ensure the chain rule of fractional-order matrix differentiation, the size of the fractional-order differentiation matrix of $\mathbf{F}(\mathbf{X})$ w.r.t. $\mathbf{X}$ is also $m\times p$. Combining the fractional-order differentiation \cref{eq12,eqC1,eqC2}, the number of different fractional-order differentiation matrices is

\begin{equation}
    \label{eqC3}
    \frac{p\times n\times m\times n}{m\times p}=n^2
\end{equation}
\end{proof}

\section{Derivation of the Fractional-order Symbolic Differentiation Formula in Artificial Neural Networks} \label{appendix4}
\renewcommand{\thefigure}{D.\arabic{figure}}
\setcounter{figure}{0}
\renewcommand{\theequation}{D.\arabic{equation}} 
\setcounter{equation}{0} 
\begin{proof}
Let $y = xw + b=xw + bw^0$. Since the fractional-order differentiation operator is a linear operator, we have
{\scriptsize
\begin{equation}
    \label{eqD1}
    \begin{aligned}
        \frac{\partial^\alpha y}{\partial w^\alpha}&={}_0^R\mathscr{D}_w^\alpha y=x\frac{\Gamma(1 + 1)}{\Gamma(1 + 1-\alpha)}|w - 0|^{1 - \alpha}+b\frac{\Gamma(0 + 1)}{\Gamma(0 + 1-\alpha)}|w - 0|^{0 - \alpha}
    \end{aligned}
\end{equation}}

Because $\Gamma(1)=\Gamma(2) = 1$, for \cref{eqD1}, we have
{\footnotesize
\begin{equation}
    \label{eqD2}
    \begin{aligned}
        \frac{\partial^\alpha y}{\partial w^\alpha}&=\frac{x}{\Gamma(2 - \alpha)}|w|^{1 - \alpha}+\frac{b}{\Gamma(1 - \alpha)}|w|^{-\alpha}
    \end{aligned}
\end{equation}}

In engineering applications, considering the need to retain the gradient direction, the sign function needs to be added to \cref{eqD2} in even-powered terms to ensure the correctness of the differentiation direction. Finally, \cref{eqD2} can be transformed into \cref{eq12}.
\end{proof}
\end{appendices}

% \bibliographystyle{unsrt}  
% \bibliography{references}  

\end{document}